\documentclass[journal,12pt,onecolumn]{IEEEtran}

\usepackage{graphicx}
\usepackage{amsmath}
\usepackage{amssymb}
\usepackage{diagbox}

\DeclareGraphicsExtensions{.pdf,.jpeg,.png}
\usepackage{subfigure}

\begin{document}

\title{Global Weighted Average Pooling Bridges Pixel-level Localization and Image-level Classification}

\author{Suo Qiu
\thanks{S. Qiu is with School of Electronic and Information Engineering, South China University of Technology, Wushan RD., Tianhe District, Guangzhou, P.R.China (e-mail: q.suo@foxmail.com)}}


\maketitle

\begin{abstract}
   In this work, we first tackle the problem of simultaneous pixel-level localization and image-level classification with only image-level labels for fully convolutional network training. We investigate the global pooling method which plays a vital role in this task. Classical global max pooling and average pooling methods are hard to indicate the precise regions of objects. Therefore, we revisit the global weighted average pooling (GWAP) method for this task and propose the class-agnostic GWAP module and the class-specific GWAP module in this paper. We evaluate the classification and pixel-level localization ability on the ILSVRC benchmark dataset. Experimental results show that the proposed GWAP module can better capture the regions of the foreground objects. We further explore the knowledge transfer between the image classification task and the region-based object detection task. We propose a multi-task framework that combines our class-specific GWAP module with R-FCN. The framework is trained with few ground truth bounding boxes and large-scale image-level labels. We evaluate this framework on PASCAL VOC dataset. Experimental results show that this framework can use the data with only image-level labels to improve the generalization of the object detection model. 
\end{abstract}



\section{Introduction}

Over the last few years supervised convolutional neural network (CNN) methods have been extremely successful for the whole-image classification \cite{He2015}, the region-based object detection \cite{ren15fasterrcnn} and the semantic image segmentation \cite{long2015fully} tasks. There are a lot of works focused on carefully designing models to improve the performance of these tasks independently. However, the drawback of such discrete supervised learning systems is data hunger. In the fully supervised setting, the performance of a learned model highly depends on the amount of training data. In order to achieve satisfactory generalization performance, a large number of precise training pairs, where each sample is associated with a label or target, are often required. Annotating training samples is often time-consuming and costly. Especially for the region-based object detection and the semantic image segmentation tasks, it is often tedious to specify exactly where the objects are and to annotate precise contours of the objects. In this paper, we explore the methods to reduce annotation costs for object localization and detection.

One solution for the reduction of annotation costs is the \emph{weakly supervised learning} (WSL) method. In the WSL setting, only image-level labels indicating the presence or absence of objects are required. WSL leave out the effort of annotating the specific position of an object and thus reduce the costs. WSL is a valuable setup for many practical applications (i.e. object localization/detection) since the weak supervision (i.e. image-level labels) is easier to obtain than the full supervision (i.e. bounding boxes). WSL is also feasible because of the inherent correlation between the whole-image classification and the object localization/detection. For example, an image was annotated as "bird" because there is exactly a bird in this image. Even though it is not sure where the bird is, information from the region of the bird is more discriminative than the uninformative backgrounds for the classification of the whole image. Various WSL methods \cite{wu2015deep,gmp,gap} demonstrate their potential of auto-localization to reduce the annotation effort.  

One branch of WSL for object detection is the region-based methods~\cite{song2014learning,bilen2015weakly,wang2015large,Bilen_2016_CVPR}. These methods first use object proposal methods (e.g. selective search \cite{ss}) to generate candidate detection results, then pick up the contributive ones to the image classification task as the confident detection results. CNN model for these methods usually contains the region-based pooling module and generates features, classifies the feature vectors for the corresponding regions. The other branch of WSL for object localization is the pixel-level localization methods~\cite{gmp,gap}. Different from the former one, CNN models for these methods do not use object proposals as the candidates, but try to predict each position on the feature map. These methods are built upon the fully convolutional architecture\footnote{Only the last layer is fully-connected.}, which is also an important class of network structures in CNN. In this paper, we mainly explore the second branch. 

\begin{figure}
	\centering
	\includegraphics[width=0.55\linewidth]{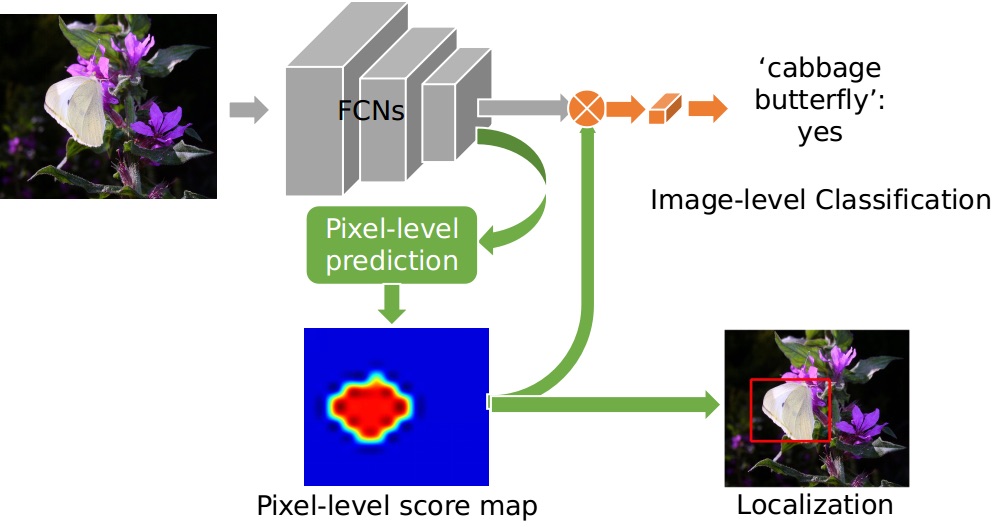}	
	\caption{Illustration of the weakly supervised learning framework for simultaneous pixel-level localization and image-level classification with fully convolutional networks. }
	\label{fig:fig1}
\end{figure} 

\begin{figure}
	\centering
	\subfigure[GMP]{ \label{fig:subfig:gmp} 
		\includegraphics[width=0.2\linewidth]{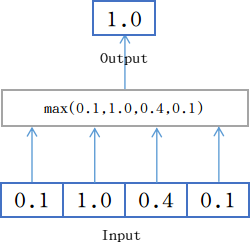}}
	\subfigure[GAP]{ \label{fig:subfig:gap} 
		\includegraphics[width=0.2\linewidth]{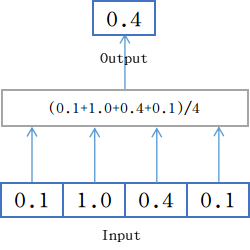}}
	\subfigure[GWAP]{ \label{fig:subfig:gwap} 
		\includegraphics[width=0.2\linewidth]{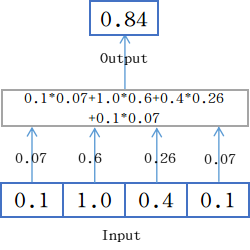}}	
	\caption{Illustration of the global pooling methods: (a) global max pooling (GMP), (b) global average pooling (GWAP), (c) global weighted average pooling (GWAP).}
	\label{fig:pool}
\end{figure}

There are two classical WSL methods, namely global max pooling (GMP) ~\cite{gmp} and global average pooling (GAP) ~\cite{gap}, for object localization. They both solve the WSL problem by the multiple instance learning (MIL) process, where pixel-level predictions are grouped into the image-level prediction and a label is attached to the whole image and not to every pixel-level prediction. In such frameworks, GMP and GAP correspond to different grouping strategies. The two strategies yield great potential for the WSL task, but both of them are hardwired. As illustrated in Fig.~\ref{fig:pool}, GMP only selects the largest value as the final result, which is compact to meet the assumption that at least one candidate prediction is the true prediction but also loses the other useful information for accurate localization. GAP is the opposite. It equally aggregates all input values as the final output, which favors the saliency input but cannot directly indicate which inputs are the true prediction. Thus, both GMP and GAP can only obtain the approximate pixel-level localization result. To this end, we revisit the global weighted average pooling (GWAP) operation for the WSL localization task. 

GWAP compute the weighted mean of the input values with their corresponding weights. These weights are non-negative and indicate the relative importance of the input values. In GWAP, the values with a higher weight contribute more to the final result than the values with a lower weight. For the WSL of object localization, the weights in GWAP naturally provide the pixel-level localization. Because the regions of objects are usually more discriminative than the backgrounds' and thus have higher weights. This property provides a way to specify the precise regions of the objects. 

To implement this idea, we propose the pixel-level prediction module to generate weights for GWAP. Fig.~\ref{fig:fig1} illustrate a basic framework for WSL with GWAP. Specifically, the pixel-level prediction module consists of a prediction unit and a normalization unit. In this work, we both design the class-agnostic prediction module and the class-specific module. Different from GMP and GAP, our GWAP method is a learnable method. This learnable weighted average operation generalizes max and average operation and also provides the capacity for more accurate localization. The proposed method is fully differentiable and easy to be trained by the standard backpropagation. 
We empirically show that this pixel-level prediction module trained only with image-level labels can effectively highlight the foreground regions and suppress backgrounds (Fig.~\ref{fig:maps}). Based on this observation, we further explore the transferability of this framework. We combine the popular fully convolutional object detection framework R-FCN~\cite{dai2016r} with our method to obtain a novel semi and weakly supervised detection method. This forms a multi-task learning framework. When the number of training samples for the object detection task is small, the WSL with GWAP can provide useful regularization to avoid overfitting and thus improve the generalization. Experimental results on the ILSVRC~\cite{ILSVRC15} and PASCAL VOC datasets~\cite{pascal-voc-2007} show the effectiveness of the proposed methods. 

\section{Related work}

In this section, we briefly review the related work on weakly supervised object detection/localization and knowledge transfer for object detection with minimal supervision. 

\subsection{Weakly supervised object detection}
Weakly supervised object detection methods~\cite{kumar2010self,deselaers2012weakly,7420739,song2014learning,bilen2015weakly,wang2015large,Bilen_2016_CVPR} are evolved from the region-based object detection frameworks~\cite{felzenszwalb2010object,girshick2016region,He2014Spatial,Girshick2015Fast}. 
They generate proposals as the candidate detections and alternately learn to find the confident proposals. Early works are formulated by shallow models and mainly focus on solving the non-convex optimization problem by using good initialization strategies \cite{deselaers2012weakly,kumar2010self}, learning strategies \cite{7420739,wang2015large} and developing smoothed models \cite{song2014learning,bilen2015weakly}. They fail to share deep representation learning and their performance is still far from the fully supervised methods.
The first deep learning framework~\cite{Bilen_2016_CVPR} for this task aggregates proposal classification scores and localization scores to image-level predictions. 
Based on this framework, context-aware network~\cite{kantorov2016contextlocnet}, expectation-maximization algorithm~\cite{Yan2017Weakly}, online instance classifier refinement process~\cite{Tang2017Multiple}
and collaborative learning method~\cite{Wang2018Collaborative} are proposed to improve the performance. These methods mainly focus on developing strategies to increase the number of proposals that really contain complete objects. 
Our work differs these methods in that, we focus on the weakly supervised pixel-level localization problem. We do not use proposals for the WSL. And our work builds upon fully convolutional networks.

\subsection{Weakly supervised object localization}

Different from these weakly supervised object detection methods, weakly supervised object localization methods build upon fully convolutional networks and output coarse regions of objects instead of the tight bounding boxes. Some works~\cite{self-taught, bency2016weakly} for this task heuristically search for regions that have important impacts on the classification. Some works~\cite{zeiler2014visualizing, simonyan2013deep} use visualization techniques to find the saliency regions for classification. Oquab et al. \cite{gmp} first unified the image-level classification and the pixel-level localization tasks into an end-to-end learning framework by using the global max pooling (GMP) method. Sun et al.~\cite{sun2016pronet} further extends the method into a multi-scale cascaded neural network and use the log-sum-exp (LSE) pooling method to improve the localization accuracies. Zhou et al. \cite{gap} revisited the global average pooling (GAP) method and proposed the class activation mapping for localizing discriminative regions. These methods have shown that the convolutional units of CNN actually behave as meaningful pattern detectors. So the location information of target objects emerges in the activation maps after convolution. These works \cite{gmp, gap, sun2016pronet} can predict approximate locations of objects and output accurate image-level labels. Even though they yield promising results, GMP and GAP operation are hardwired. LSE uses a hyper parameter to adjust the pooling scale but is also not flexible enough. Durand et al. \cite{Durand_2016_CVPR,durand2017wildcat} proposed methods to select multiple high and low scoring regions instead of a single highest scoring region. However, their methods also need to fine-tune the hyperparameters for selection and are also restrained. These works demonstrate that the regions of target objects have more effects on the whole-image classification and inspire us to explore the localization ability of fully convolutional networks. To obtain more precise localization results, we revisit the global weight average pooling (GWAP) operation and propose the class-agnostic and class-specific GWAP modules for this task. Different from prior works, the proposed modules are learnable and flexible to capture various appearances of objects. In addition, we first explore the combination between the weakly supervised module and the region-based detection framework in this paper. 

\subsection{Knowledge transfer for object detection with minimal supervision}

Despite using the image-level labels, transferring knowledge from an auxiliary dataset also can improve the learning performances when lack of supervised data for object detection. Hoffman et al.~\cite{hoffman2014lsda} use labeled data to train the transfer model between image classification and object detection, then use this transfer model for unlabeled data. Shi et al.~\cite{shi2017transfer} uses a ranking model to enhance the selection of the detection results. Guillaumin et al.~\cite{guillaumin2012large} transfer the knowledge about the plausible location, appearance, and context of the target objects from the similar classes. Chen et al.~\cite{chen2018lstd} design a flexible deep architecture, a background depression regularization and a transfer-knowledge regularization to alleviate transfer difficulties. Different from these works, our work does not use extra datasets. And we transfer knowledge between the weakly supervised object localization task and the supervised object detection task.

\section{Proposed Method}
\label{sec:model}

\subsection{Global Weighted Average Pooling for WSL}
\label{sec:gwap}

Simultaneous pixel-level localization and image-level classification with only image-level labels is a weakly supervised learning (WSL) problem. In this subsection, we discuss how the global weighted average pooling method can be used for this task.

In this work, we build our method based on the fully convolutional network (FCN). In convolutional neural networks, the convolution operation actually can be seen as a pattern detection process in a sliding window manner. And each activation after the convolution along the spatial dimension can be used to specify whether the pattern is located at this position or not. These activations reflect the pixel-level localizations of objects in an image and also indicate the importance of corresponding features for recognizing the objects. The pixel-level information and the image-level recognition are correlative. For example, if the image-level result says that there is a cat, some pixel-level predictions must indicate the cat too. If the image-level result says that there is no cat, any pixel-level predictions should not be the cat. Thus, the entire image-level recognition can be obtained by aggregating the pixel-level results. 

In this paper, we use the global weighted average pooling (GWAP) method to bridge the pixel-level localization and the image-level classification in fully convolutional networks. GWAP can be used in two ways. One is aggragating features. The other is aggragating scores. At the spatial location $(x,y)$, let $s_c(x,y)$ represents the prediction score of class $c$ in the classification convolutional layer and $f_k(x,y)$ represents the activation of unit $k$ in the last convolutional layer before the classification layer, respectively. For aggragating scores, the image-level prediction 
\begin{equation}
\label{equ:gwaps}
\begin{split}
S_c&=\sum_{x,y}\alpha(x,y)s_c(x,y),
\end{split}
\end{equation}
where $\alpha(x,y)$ is the weight at the spatial location $(x,y)$. For aggragating features, the image-level prediction
\begin{equation}
\label{equ:gwapf} 
S_c=\sum_{k}\omega_{k}^{c}\sum_{x,y}\alpha(x,y)f_k(x,y),
\end{equation}
where $\omega_{k}^{c}$ is the classification weight corresponding to class $c$ for unit $k$.
When $s_c(x,y)=\sum_{k}\omega_{k}^{c}f_k(x,y)$, Equ.~\eqref{equ:gwaps} and Equ.~\eqref{equ:gwapf} are equivalent. 

In our task, the weight $\alpha(x,y)$ should satisfy the following principles. First, the weights should be non-negative and some may be zero, but not all of them. That is $\forall \alpha(x,y) \geq 0$ and $\sum_{x,y}\alpha(x,y)\neq0$. Naturally, a higher weight means the more contribution to the aggregating result than a lower one. Second, the weights should be normalized such that they sum up to $1$. That is $\sum_{x,y}\alpha(x,y)=1$. With these properties, GWAP performs the gating function. The model with GWAP passes scores/features with high weights and suppress the scores/features with low weights. Therefore, the weights $\alpha$ should be able to indicate the object regions which are informative for the image-level classification. 

In this paper, we design the \emph{pixel-level prediction module} to generate $\alpha$. With the above analysis, the pixel-level prediction module contains the pixel-level classification unit and the spatial normalization unit. The pixel-level classification unit outputs predictions of the pixel-level locations. The spatial normalization unit normalizes these predictions for aggregating. In this work, we design both the class-agnostic (Section~\ref{sec:ca}) and the class-specific (Section~\ref{sec:cs}) pixel-level prediction modules.

\subsection{Class-agnostic Pixel-level Prediction}
\label{sec:ca}
\begin{figure}[!t]
	\centering
	\includegraphics[width=0.6\linewidth]{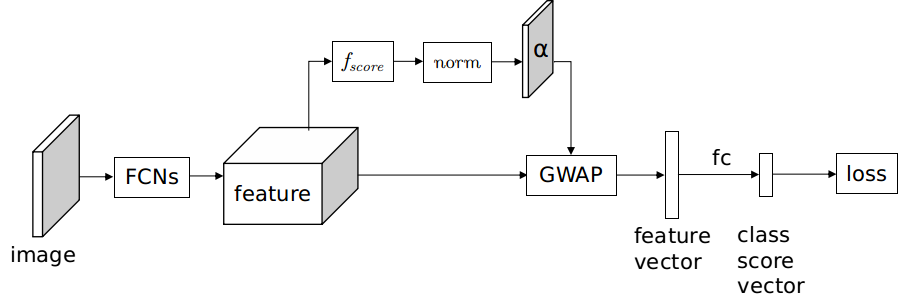}	
	\caption{The network architecture of the proposed class-agnostic GWAP method for image classification. ($f_{score}$: Equ.~\ref{equ:cascore}, norm: Equ.~\ref{equ:canorm}, GWAP: Equ.~\ref{equ:gwap2}, fc: fully conneted layer)}
	\label{fig:canet}
\end{figure}

In subsection~\ref{sec:gwap}, we introduce the intuition of using global weighted average pooling for simultaneous pixel-level localization and image-level classification. In this section, we introduce the class-agnostic implementation for GWAP. 

In the class-agnostic implementation, there is only one weight map $\alpha$ for the classification. We use this shared weight map to aggregate features and then classify the aggregated feture vector. Fig.~\ref{fig:canet} illustrates the network architecture of the poposed method. Concretely, FCNs caculate the features of the input image. We obtain the feature maps $\mathbf{F} \in R^{K \times H \times W}$ from the last convolutional layer. A bypass network is added after $\mathbf{F}$ to generate the weight map $\alpha$. The weight map $\alpha$ is computed by the following functions 
\begin{equation}
\label{equ:cascore}
\begin{split}
\mathbf{M}(x,y)&=f_{score}(\mathbf{F}(x,y)) \\
&=\exp(\sigma(\mathbf{w}\mathbf{F}(x,y)+\mathbf{b})),
\end{split}
\end{equation}
\begin{equation}
\label{equ:canorm}
\alpha(x,y)=\frac{\mathbf{M}(x,y)}{\sum_{x,y}\mathbf{M}(x,y)},
\end{equation}
where $1 \leq x \leq W$, $1 \leq y \leq H$. And $K$, $H$ and $W$ denote the number of channels, the height of the feature maps and the width of the feature maps, respectively. $\mathbf{w} \in R^{1 \times K}$ is the parameter matrix and $\mathbf{b} \in R^{1 \times 1}$ is the corresponding bias. $\sigma(\cdot)$ is the \emph{sigmoid} function. $\exp(\cdot)$ is the \emph{exponentiate} operation. Once the weight map $\alpha$ is computed, the aggregated feature vector of the whole image is computed by
\begin{equation}
\label{equ:gwap2}
\mathbf{f}=\sum_{x,y}\alpha(x,y)\mathbf{F}(x,y), \mathbf{f} \in R^{K \times 1}.
\end{equation}
Then, one fully connected layer is used to classify the aggregated feature vector $\mathbf{f}$ and the task-specific loss function is used to train the whole network. After training with this framework, the class-agnostic pixel-level prediction module can distinguish foreground objects and backgrounds. The proposed network ouputs accurate image-level labels and predicts precise regions of objects simultaneously in a forward pass.

\textbf{Comparision with the soft attention model:} 
Soft attention model (SAM) is proposed by Bahdanau \cite{sam} in neural machine translation area, and has been successfully used for computer vision tasks \cite{xu2015show,sharma2015action,Chen_2016_CVPR,Yang_2016_CVPR,You_2016_CVPR,Lee_2016_CVPR,li2016attentive}. It searches for parts of the input that are relevant to the final prediction without any advance annotations. This has the same spirit with our task. However, SAM did not work very well for our task in our early experiment. The score function of SAM is just an exponentiate operation ($\exp$), which is just an approximation of the $\max$ operation. Thus SAM is also hard to obtain precise object regions. Different from SAM, we use the sigmoid function to provide a non-linear bound for the prediction. This makes it easier to capture the various appearances of an object and thus has a better localization property. We verify this experimentally in Section~\ref{exp:ce}.

\subsection{Class-specific Pixel-level Prediction}
\label{sec:cs}

In some applications, we might need to know the class of a highlight region in the pixel-level prediction module. Thus, we further propose the class-specific pixel-level prediction module. In Section~\ref{sec:swd}, we use this module for semi and weakly supervised detection.

\begin{figure}[!t]
	\centering
	\includegraphics[width=0.8\linewidth]{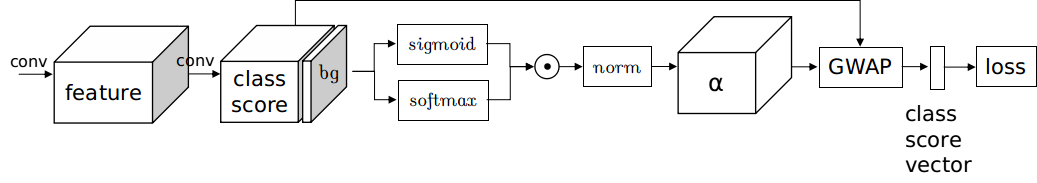}	
	\caption{The network architecture of the proposed class-specific GWAP module for image classification. (bg: background class, $\odot$: element-wise product, norm: Equ.~\ref{equ:csnorm}, $\alpha$: class-specific weight map, GWAP: Equ.~\ref{equ:csgwap}.)}
	\label{fig:csnet}
\end{figure} 

Fig.~\ref{fig:csnet} illustrates the design of the proposed class-specific GWAP module. First, we add a convolution layer to obtain the class score map $\mathbf{S} \in R^{C \times H \times W}$, where $C$ is the number of object classes, $H$ and $W$ are the height and the width of the score maps. We also add one more channel to encode backgrounds $\mathbf{B} \in R^{1 \times H \times W}$. We concat $\mathbf{S}$ and $\mathbf{B}$ and obtain $\mathbf{\hat{S}} \in R^{(C+1) \times H \times W}$. To constrain that every position on the map can only belong to a unique class, we use the softmax operation, defined as follows:
\begin{equation}
\label{equ:cssoftmax}
\mathbf{O}_c(x,y)=\frac{\exp[\mathbf{\hat{S}}_c(x,y)]}{\sum_{c=1}^C\exp[\mathbf{\hat{S}}_c(x,y)]}, \forall c\in[1,C+1],
\end{equation}
where $1 \leq x \leq W$, $1 \leq y \leq H$. After obtaining $\mathbf{O_c}$, the final score map 
\begin{equation}
\label{equ:csscore}
\mathbf{M}_c = \mathbf{O}_c \odot \sigma(\mathbf{S}_c), \forall c\in[1,C],
\end{equation}
where $\sigma(\cdot)$ is the sigmoid function, and $\odot$ is the element-wise product. Then the class-specific weight maps 
\begin{equation}
\label{equ:csnorm}
\alpha_c(x,y)=\frac{\exp[\mathbf{M}_c(x,y)]}{\sum_{x,y}\exp[\mathbf{M}_c(x,y)]}, \forall c\in[1,C].
\end{equation}
Finally, we obtain the aggregated score vector for the whole image. That is 
\begin{equation}
\label{equ:csgwap}
\mathbf{s} = \sum_{x,y}\alpha(x,y) \odot \mathbf{S}(x,y), \mathbf{s} \in R^{C \times 1}.
\end{equation}
The class-specific module outputs localization for each class as shown in Fig.~\ref{sec:swd}. The class-specific and the class-agnostic GWAP module have the same design ideas and thus the same properties. They both learnable and easy to train. They all can predicts precise regions of objects.

\subsection{Semi and Weakly Supervised Detection}
\label{sec:swd}

\begin{figure}[!t]
	\centering
	\includegraphics[width=0.8\linewidth]{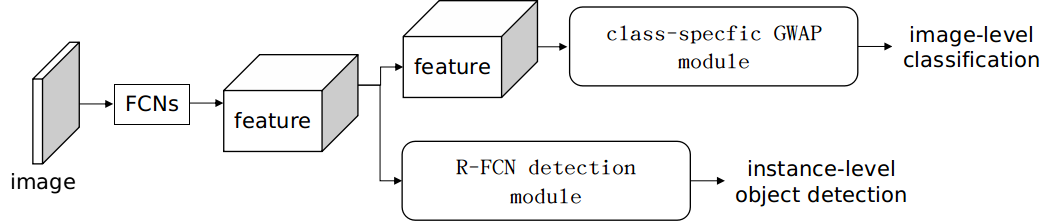}	
	\caption{Illustration of the multi task framework for semi and weakly supervised detection.}
	\label{fig:rfcngwap}
\end{figure}

\begin{figure}[!t]
	\centering
	\includegraphics[width=0.6\linewidth]{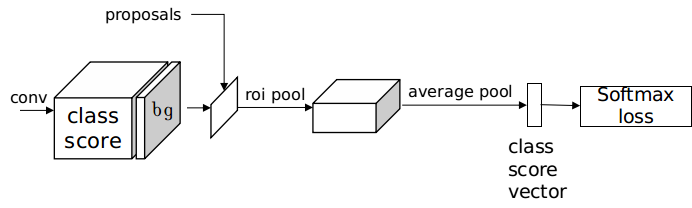}	
	\caption{Illustration of the local region regularization for the class-specific score map.}
	\label{fig:lr}
\end{figure}

In order to further explore whether the GWAP module can share features with the object detection task, we propose the multi-task learning framework. As illustrated in Fig.~\ref{fig:rfcngwap}, this framework contains two branches. One is the standard object detection pipeline. In this work, we choose the R-FCN~\cite{dai2016r} model as the detection branch. Because R-FCN is also fully convolutional, which is compatible with our GWAP module. The R-FCN branch can only be trained with images that have ground truth bounding boxes. The other branch is the image classification branch with GWAP module. Here we use the class-specific GWAP module mentioned in Section~\ref{sec:cs} for the image classification branch. This branch can be trained with only image-level labels. For better performance, we further add local object region regularization for the class-specific score map (illustrated in Fig.~\ref{fig:lr}). With the ground truth bounding boxes, we constraint that the average score of the object region is larger than the negative samples. In this work, we train the object detection task and the image classification task jointly. The total loss is $loss = loss_{det} + \lambda loss_{weak}$. $\lambda = 0.1$ in our work. $\lambda$ balances the two tasks to avoid overfitting to one of them. Experimental results show that this multi-task framework improves the generalization when lack of supervised training data for detection.  

\section{Ablation experiments}
\label{exp:ce}

In this section, we first conduct some ablation experiments to illustrate the effectiveness of the design of our class-agnostic GWAP method (Section~\ref{sec:ca}). Without loss of generality, we only perform experiments with the standard end-to-end classification process, which is much simpler than a specialized model for the task. The observation conclusions here are also applicable to the class-specific GWAP method because the class-agnostic and class-specific GWAP methods have the same design ideas.

\subsection{Setup}
We conduct a set of basic experiments to understand how GWAP compares to GAP \cite{gap} and evaluate the design of the class-agnostic GWAP method. For fair comparison, we implement these methods on the same basic network architecture, such as the CaffeNet (essentially AlexNet \cite{alexnet}) and GoogLeNet \cite{googlenet}. We call CaffeNet model \textbf{C} and GoogLeNet model \textbf{G}, respectively. We use the pre-trained ImageNet model that are available online.\footnote{https://github.com/rbgirshick/fast-rcnn} \footnote{https://github.com/BVLC/caffe/wiki/Model-Zoo} For CaffeNet, we remove the layers after $conv5$ and replace them with GWAP module or GAP followed by a classification layer. For GoogLeNet, we remove the layers after $inception4e$ (i.e., pool4 to prob). For multi-label classification, we use sigmoid cross entropy loss in this section. Then we fine-tune these networks on the trainval set of PASCAL VOC2007 \cite{pascal-voc-2007} and evaluate it on the test set. All experiments use the single-scale training and testing ($600\times600$ input image). During training, each sampled image is horizontally flipped. No other data augmentation is used. The size of mini-batch is set to 4 images. This is a relatively small batch size. We set the initial learning rate to $0.0005$, and decrease it every $20,000$ iterations. We run a total of $60,000$ iterations.

\subsection{Results}

\begin{table*}[t] \small 
	\caption{VOC 2007 test classification average precision ($\%$) of the ablation experiments (C: CaffeNet, G: GoogLeNet, w/o: without, $\sigma$: sigmoid function, $\exp$ exponentiate operation, -gt: using ground truth attention map).}
	\label{V07_cls}
	\centering
	\begin{tabular}{l|c|c}
		\hline \hline
		method& net 
		& \textbf{mAP} \\
		\hline
		GAP&C
		&68.6\\
		GWAP w/o $\sigma$ &C
		&57.7\\
		GWAP w/o $\exp$ &C
		&68.4\\
		GWAP&C
		&\textbf{70.4}\\
		GWAP-gt&C
		&73.7\\
		\hline
		GAP&G
		&84.7\\
		GWAP w/o $\sigma$ &G
		&81.5\\
		GWAP w/o $\exp$ &G
		&83.7\\
		GWAP&G
		&\textbf{85.1}\\
		GWAP-gt&G
		&86.5\\
		\hline
	\end{tabular}
\end{table*}

\begin{table}[t] \small
	\centering
	\caption{Attention effectiveness on the VOC 2007 test set (C: CaffeNet, G:  GoogLeNet, w/o: without, $\sigma$: sigmoid function, $\exp$ exponentiate operation).}
	\label{V07_AE}
	\begin{tabular}{l|c|c}
		\hline \hline
		method&net&F-measure\\
		\hline
		GAP&C&0.3517\\
		GWAP w/o $\sigma$&C&0.1344\\
		GWAP w/o $\exp$&C&0.3611\\
		GWAP&C&\textbf{0.5060}\\
		\hline
		GAP&G&0.4164\\
		GWAP w/o $\sigma$&G&0.1905\\
		GWAP w/o $\exp$&G&0.4577\\
		GWAP&G&\textbf{0.5450}\\
		\hline
	\end{tabular}
\end{table}

\begin{figure}[!t]
	\centering
	\begin{minipage}[b]{0.7\linewidth}
		\subfigure[Original images]{
			\includegraphics[width=0.19\linewidth]{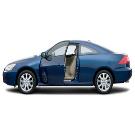}
			\includegraphics[width=0.19\linewidth]{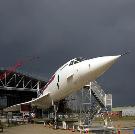}
			\includegraphics[width=0.19\linewidth]{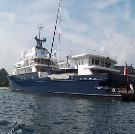}
			\includegraphics[width=0.19\linewidth]{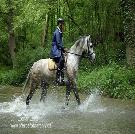}
			\includegraphics[width=0.19\linewidth]{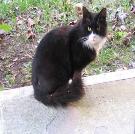}
		}
			\\
		\subfigure[GoogLeNet-GAP]{
			\includegraphics[width=0.19\linewidth]{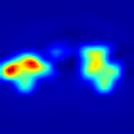}
			\includegraphics[width=0.19\linewidth]{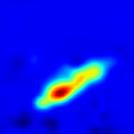}
			\includegraphics[width=0.19\linewidth]{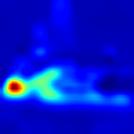}
			\includegraphics[width=0.19\linewidth]{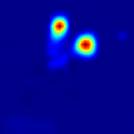}
			\includegraphics[width=0.19\linewidth]{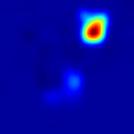}
		}
		\\
		\subfigure[GoogLeNet-GWAP without sigmoid function]{
			\includegraphics[width=0.19\linewidth]{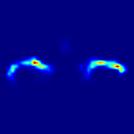}
			\includegraphics[width=0.19\linewidth]{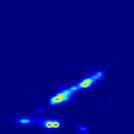}
			\includegraphics[width=0.19\linewidth]{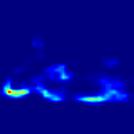}
			\includegraphics[width=0.19\linewidth]{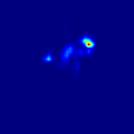}
			\includegraphics[width=0.19\linewidth]{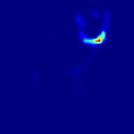}
		}
		\\
		\subfigure[GoogLeNet-GWAP without exponentiate operation]{
			\includegraphics[width=0.19\linewidth]{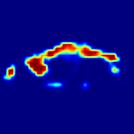}
			\includegraphics[width=0.19\linewidth]{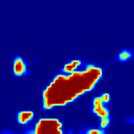}
			\includegraphics[width=0.19\linewidth]{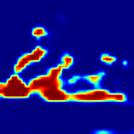}
			\includegraphics[width=0.19\linewidth]{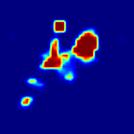}
			\includegraphics[width=0.19\linewidth]{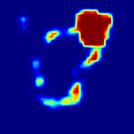}
		}
		\\
		\subfigure[GoogLeNet-GWAP]{
			\includegraphics[width=0.19\linewidth]{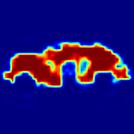}
			\includegraphics[width=0.19\linewidth]{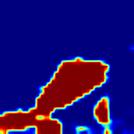}
			\includegraphics[width=0.19\linewidth]{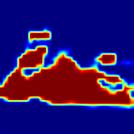}
			\includegraphics[width=0.19\linewidth]{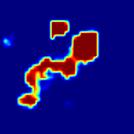}
			\includegraphics[width=0.19\linewidth]{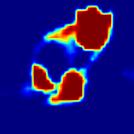}
		}
			\\
			
	\end{minipage}
	\caption{Illustration of attention maps from different models.}
	\label{fig:ae}
\end{figure}

\textbf{Classification:} We provide classification results on PASCAL VOC2007 test set (Table~\ref{V07_cls}) to demonstrate the effectiveness of the proposed method. Firstly, we verify the intuition (Sec.~\ref{sec:gwap}) of the idea of GWAP. We provide the classification results of GWAP with the ground truth weight map (refer to as GWAP-gt). Given the ground truth bounding boxes of target objects, we set values that belong to the regions of objects as 1 and the rest as 0. We normalize these values (sum to 1) to construct the ground truth weight map. As shown in Table~\ref{V07_cls}, GWAP-gt provides an improvement over GAP in mAP (CaffeNet: from 68.6$\%$ to 73.7$\%$, GoogLeNet: from 84.7$\%$ to 86.5$\%$). This demonstrates that having the regions of objects is helpful for the whole-image classification. And our method GWAP also yields an improvement over GAP (CaffeNet: from 68.6$\%$ to 70.4$\%$, GoogLeNet: from 84.7$\%$ to 85.1$\%$). We also provide the results of the class-agnostic GWAP without sigmoid function (refer to as GWAP w/o $\sigma$) and without exponentiate operation (refer to as GWAP w/o $\exp$). Results show that performance drops a lot without the sigmoid function. And without the exponentiate operation, performance also drops but better than the GWAP w/o $\sigma$. This demonstrates the importance of using the non-linear sigmoid function. In addition, only using the sigmoid function is seemingly insufficient to obtain a higher performance than GAP. The exponentiate operation also enhances the result to some extent. 

\textbf{Object localization effectiveness:} We use \emph{F-measure} to quantitatively evaluate the performance of object localization in the weight map $\alpha$, which is borrowed from salient object detection \cite{borji2015salient}. Concretely, we convert a weight map to a heat map normalized to [0,1]. Given a heat map $S$, we then convert it to a binary mask $M$ and compute \emph{Precision} and \emph{Recall} by comparing $M$ with ground-truth $G$: $Precision=\frac{|M \cap G|}{|M|}, Recall=\frac{|M \cap G|}{|G|}$. \emph{F-measure} is obtained by 
$F_\beta=\frac{(1+\beta^2)Precision \times Recall}{\beta^2Precision+Recall}$, where $\beta^2$ is set to $0.3$. 
\emph{F-measure} measures the ability to indicate the full extents of objects with minimal noise. Setting the whole region to foreground can easily lead to $100\%$ recall, but is meaningless. In this part, the ground truth regions are formed by the given bounding boxes of target objects. For an image, we aggregate regions of multi boxes into one region, for compatibility with our class-agnostic weight map. We use the adaptive threshold method Otsu \cite{otsu1975threshold} for binarizing $S$. Quantitative results are shown in Table~\ref{V07_AE} and some visualization results are given in Fig.~\ref{fig:ae}. GWAP achieves the highest score, which demonstrates the localization ability of the proposed method. We also find that there is a positive correlation between the object localization effectiveness and the performance of classification. This is consistent with our intuition.

\section{Results on ILSVRC}
\label{exp:soa}

\begin{table}[t] \small
	\caption{Classification error on the ILSVRC validation set. ($\%$)}
	\label{cls}
	\centering
	\begin{tabular}{l|c|c}
		\hline \hline
		Networks & top-1 val. error & top-5 val.error \\
		\hline
		VGGnet-GAP \cite{gap} & 33.4 & 12.2 \\
		GoogLeNet-GAP \cite{gap} & 35.0 & 13.2 \\
		AlexNet*-GAP \cite{gap} & 44.9 & 20.9 \\
		AlexNet-GAP \cite{gap} & 51.1 & 26.3 \\
		\hline
		GoogLeNet \cite{gap} & 31.9 & 11.3 \\
		VGGnet \cite{gap} & 31.2 & 11.4 \\
		AlexNet \cite{gap} & 42.6 & 19.5 \\
		NIN \cite{gap} & 41.9 & 19.6 \\
		\hline
		GoogLeNet-GMP \cite{gap} & 35.6 & 13.9 \\
		\hline
		GoogLeNet-GWAP &\textbf{31.8} &\textbf{11.3}\\
		\hline
	\end{tabular}
\end{table}

\begin{table}[t] \small
	\caption{Localization error on the ILSVRC validation set. ($\%$)}
	\label{loc}
	\centering
	\begin{tabular}{l|c}
		\hline \hline
		Networks & top-1 val. error  \\
		\hline
		GoogLeNet-GAP \cite{gap} & 56.40  \\
		VGGnet-GAP \cite{gap} & 57.20  \\
		GoogLeNet \cite{gap} & 60.09  \\		
		AlexNet*-GAP \cite{gap} & 63.75  \\
		AlexNet-GAP \cite{gap} & 67.19  \\
		NIN \cite{gap} & 65.47  \\
		\hline
		Backprop on GoogLeNet \cite{gap} & 61.31 \\
		Backprop on VGGnet \cite{gap} & 61.12 \\
		Backprop on AlexNet \cite{gap} & 65.17 \\	
		\hline
		GoogLeNet-GMP \cite{gap} & 57.78 \\
		\hline
		GoogLeNet-GWAP & \textbf{54.99} \\
		GoogLeNet-GWAP (multi-scale) & \textbf{54.09} \\
		\hline
	\end{tabular}
\end{table}

\begin{figure*}[t]
	\centering
	\subfigure{
		\centering
		\begin{minipage}[b]{0.9\linewidth}
			\centering
			\includegraphics[width=0.14\linewidth]{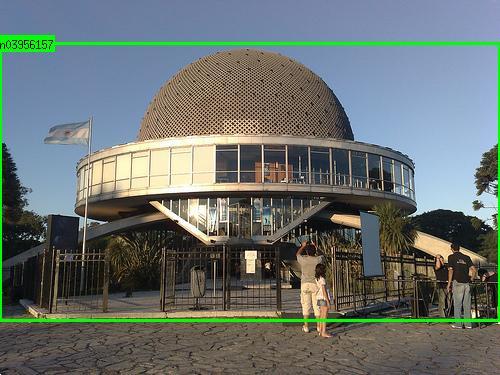} 
			\includegraphics[width=0.14\linewidth]{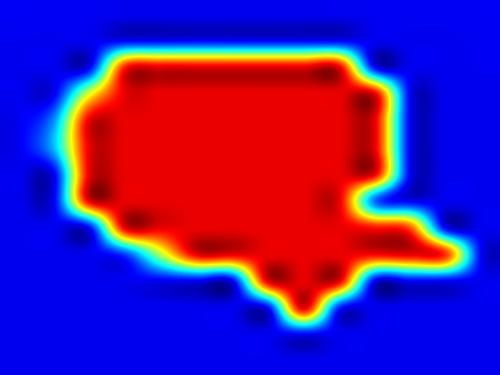} 
			\includegraphics[width=0.14\linewidth]{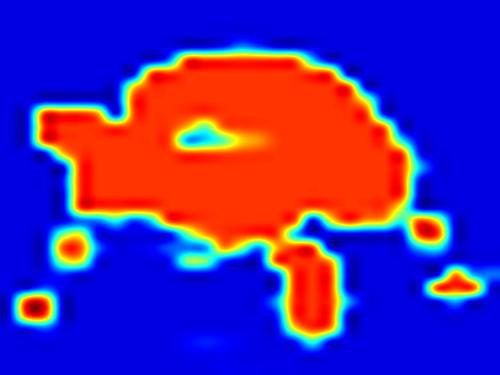} 
			\includegraphics[width=0.14\linewidth]{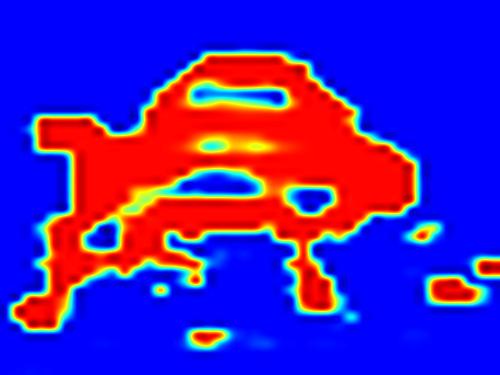} 
			\includegraphics[width=0.14\linewidth]{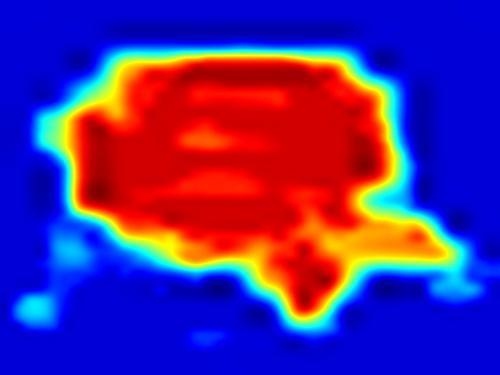} 
			\includegraphics[width=0.14\linewidth]{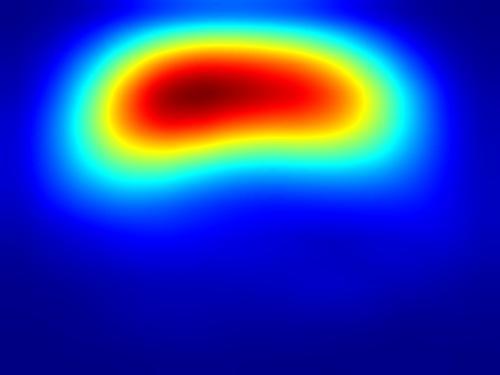}
			\\
			\includegraphics[width=0.14\linewidth]{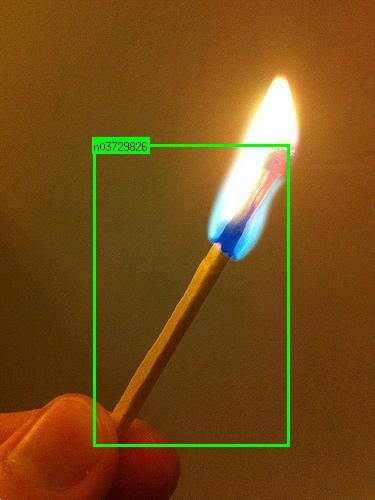}
			\includegraphics[width=0.14\linewidth]{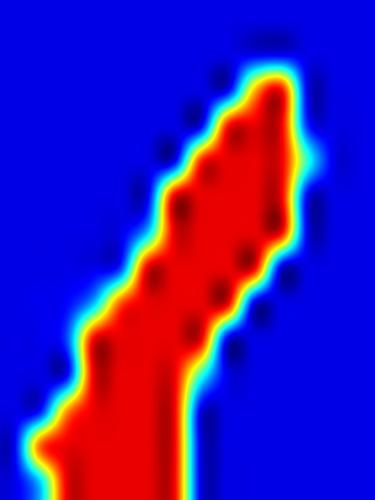}
			\includegraphics[width=0.14\linewidth]{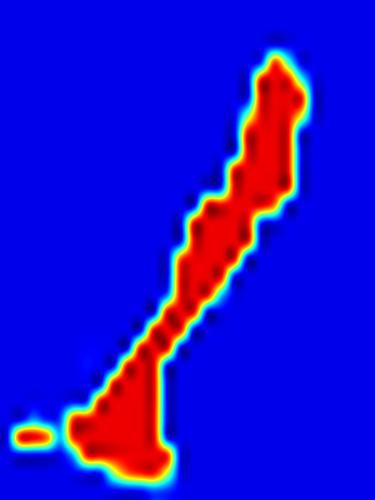}
			\includegraphics[width=0.14\linewidth]{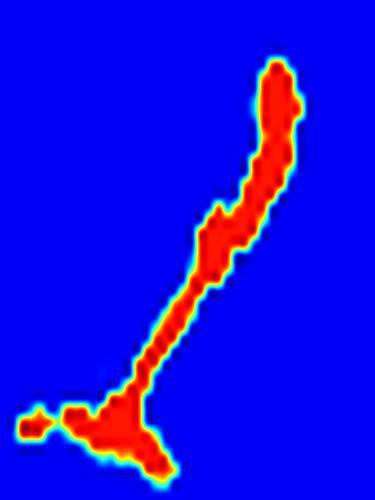}
			\includegraphics[width=0.14\linewidth]{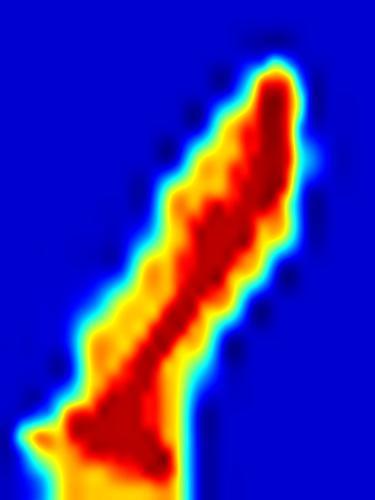}
			\includegraphics[width=0.14\linewidth]{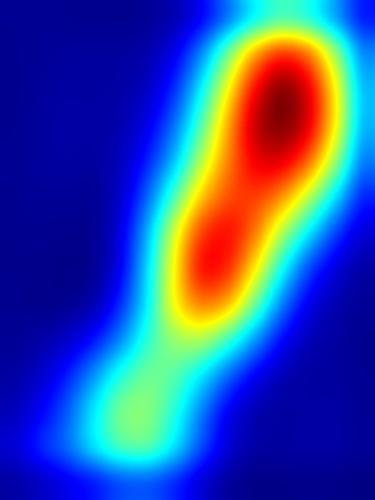}
			\\
			\includegraphics[width=0.14\linewidth]{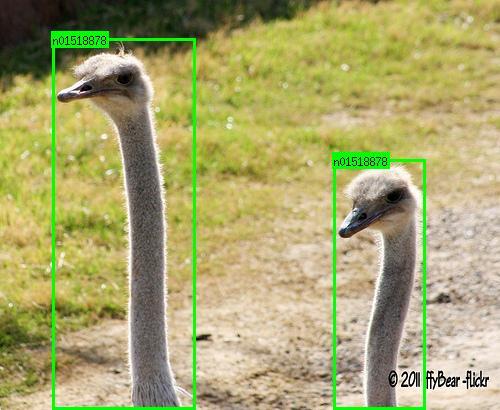}
			\includegraphics[width=0.14\linewidth]{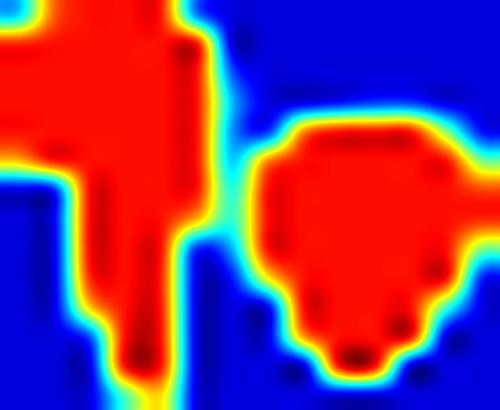}
			\includegraphics[width=0.14\linewidth]{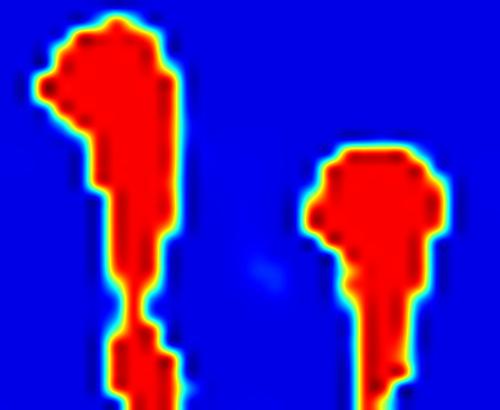}
			\includegraphics[width=0.14\linewidth]{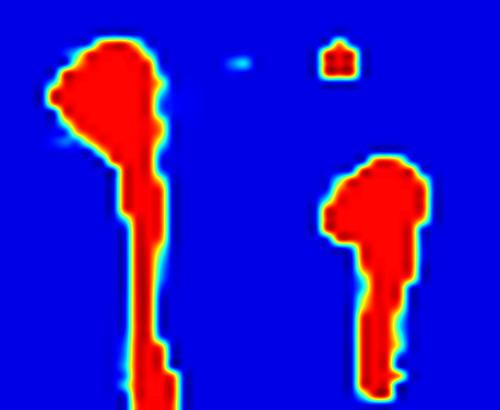}
			\includegraphics[width=0.14\linewidth]{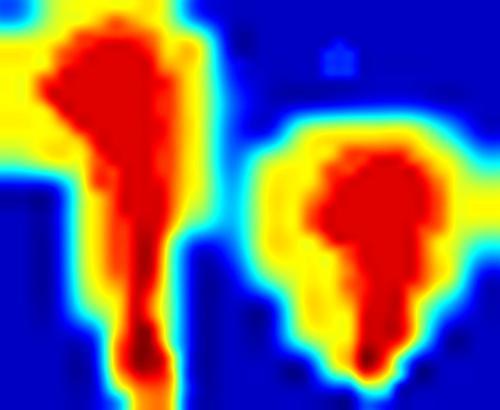}
			\includegraphics[width=0.14\linewidth]{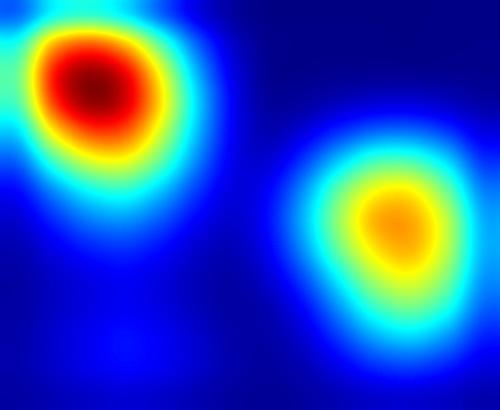}
			\\
			\includegraphics[width=0.14\linewidth]{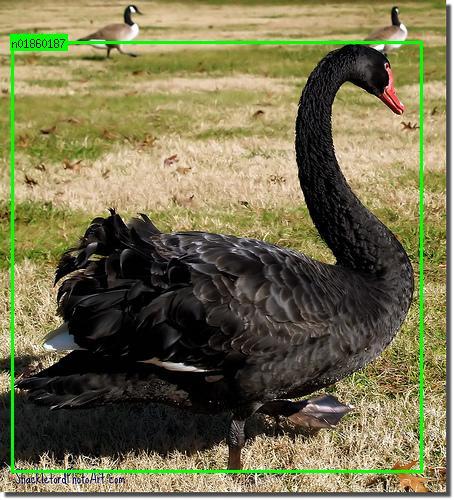}
			\includegraphics[width=0.14\linewidth]{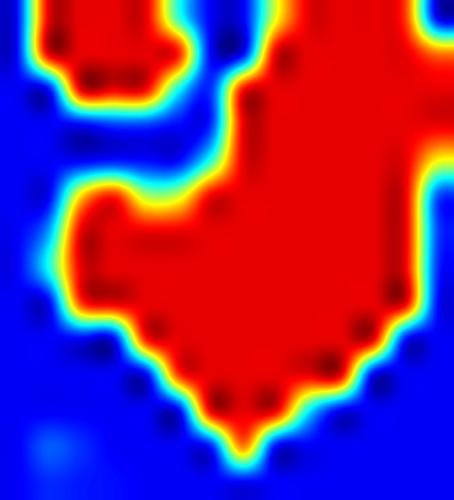}
			\includegraphics[width=0.14\linewidth]{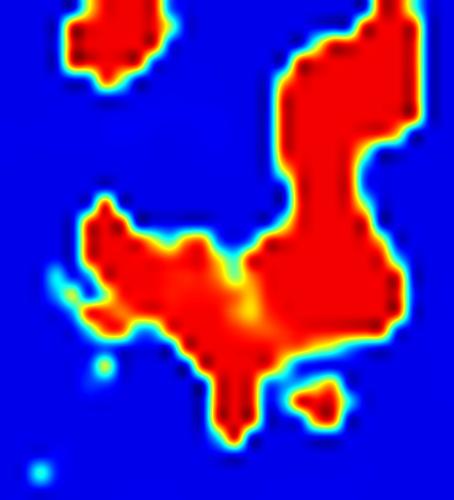}
			\includegraphics[width=0.14\linewidth]{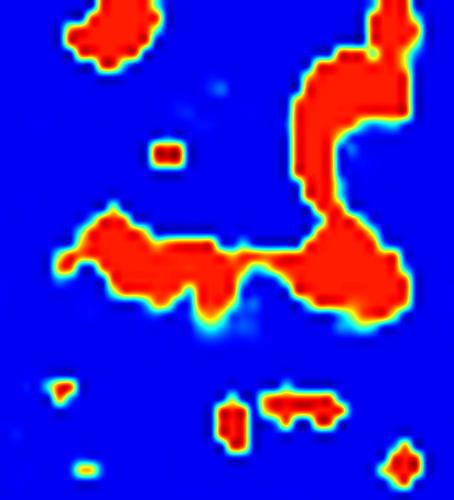}
			\includegraphics[width=0.14\linewidth]{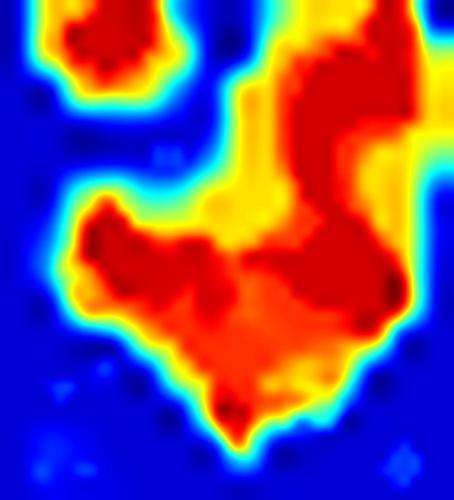}
			\includegraphics[width=0.14\linewidth]{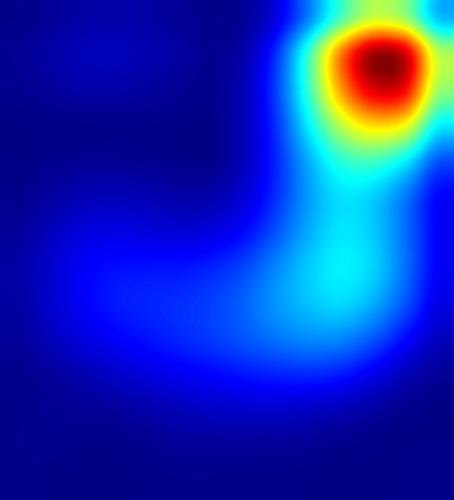}
			\\
			\includegraphics[width=0.14\linewidth]{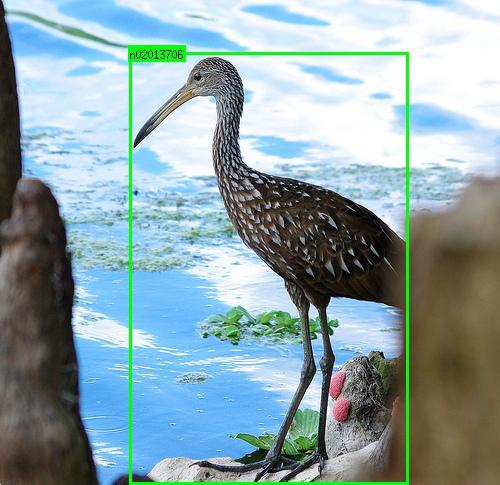}
			\includegraphics[width=0.14\linewidth]{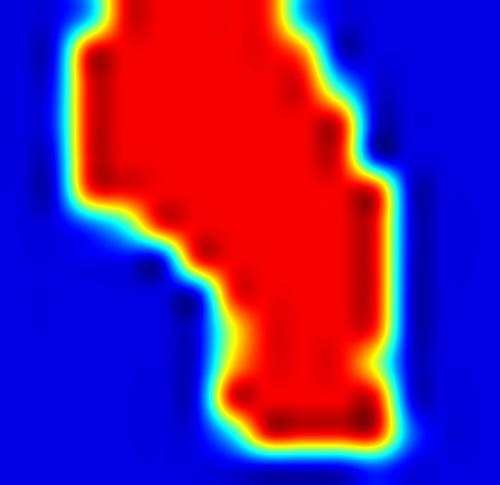}
			\includegraphics[width=0.14\linewidth]{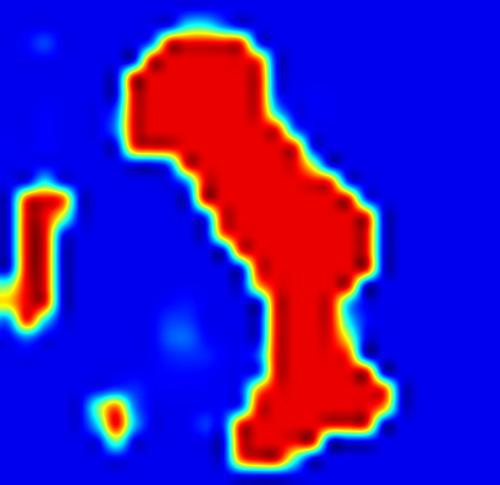}
			\includegraphics[width=0.14\linewidth]{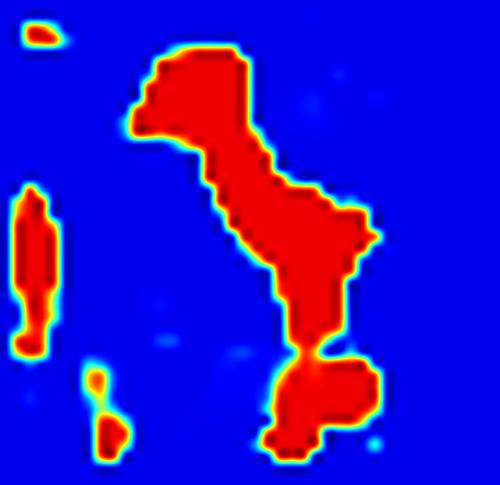}
			\includegraphics[width=0.14\linewidth]{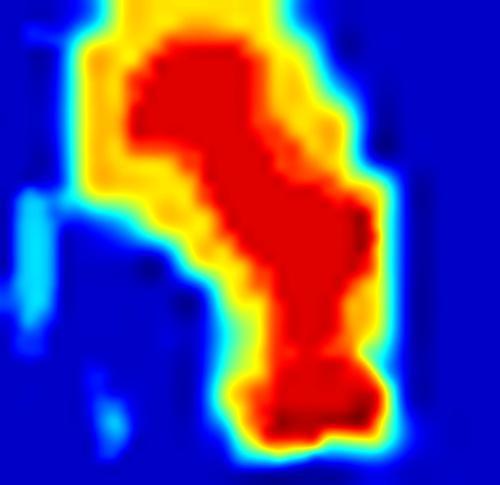}
			\includegraphics[width=0.14\linewidth]{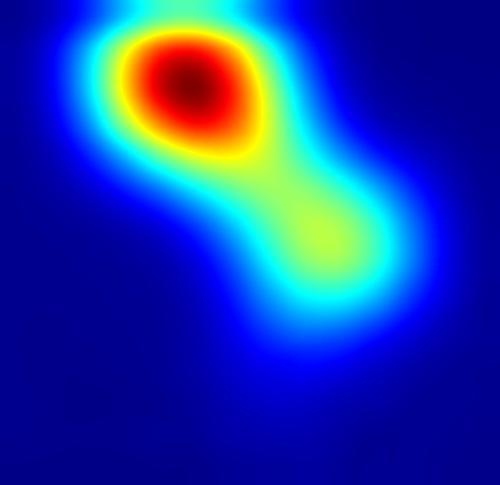}
			\\
			\includegraphics[width=0.14\linewidth]{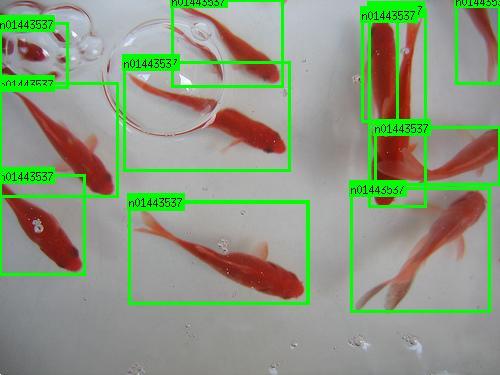}
			\includegraphics[width=0.14\linewidth]{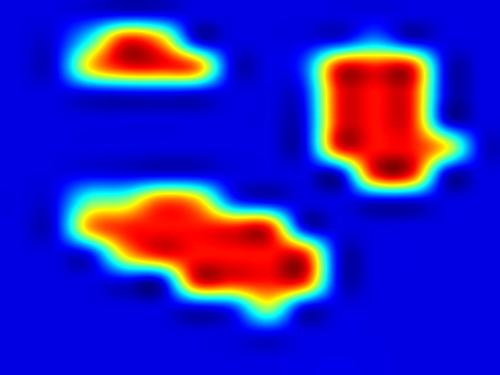}
			\includegraphics[width=0.14\linewidth]{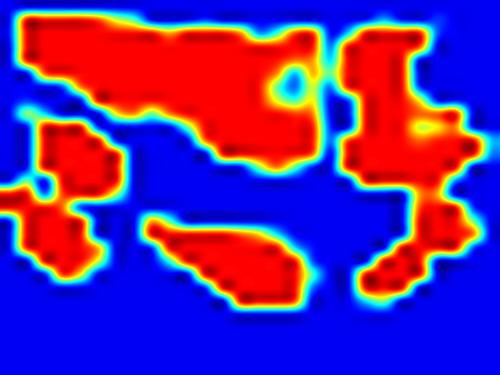}
			\includegraphics[width=0.14\linewidth]{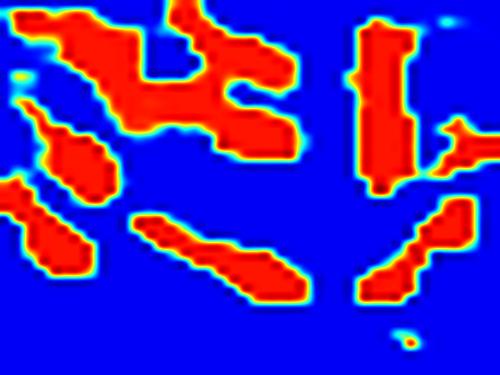}
		    \includegraphics[width=0.14\linewidth]{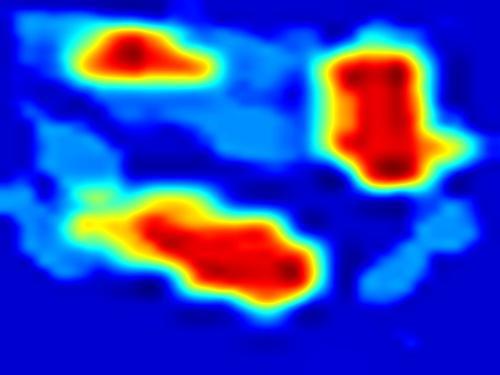}
		    \includegraphics[width=0.14\linewidth]{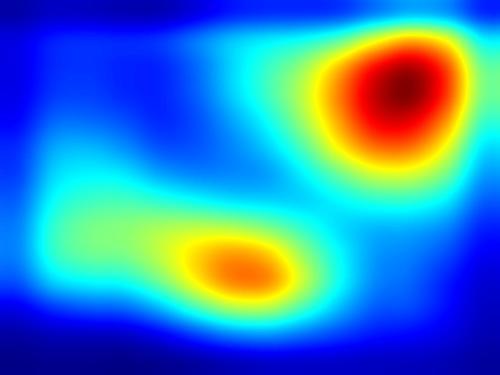}
			
		\end{minipage}
	}
	\caption{Illustration of example outputs. Fisrt column is original images with ground truth bounding boxes in green. Second, third and fourth columns are weight maps from our GoogLeNet-GWAP model with $224\times224$, $448\times448$, $672\times672$ sized input images, respectively. The average of the three scaled weight maps are shown in the fifth column. The last column is top-1 predicted class activation maps from GoogLeNet-GAP model.  }
	\label{fig:maps}
\end{figure*}

In this section, we evaluate the efficiency of the class-agnostic GWAP method on the large-scale ILSVRC 2014 benchmark dataset \cite{ILSVRC15}. We also provide both classification and localization results. We use the same experimental setup as in GAP \cite{gap}. We fine-tune the GoogLeNet \cite{googlenet} with our class-agnostic GWAP module on the pretrained imagenet model for 1000-way object classification.  

\textbf{Classification:} We use the same error metrics(top-1, top-5) as ILSVRC for classification. Table~\ref{cls} shows the classification performance of the original networks, GAP networks, and our GoogLeNet-GWAP network. We find that GoogLeNet-GWAP provides a large improvement over GoogLeNet-GAP (top-1: 3.2$\%$, top-5: 1.9$\%$), and is also comparable to the original GoogLeNet. These results demonstrate the efficiency of the proposed method. Compared to GAP, our method does not have the performance drop problem.

\textbf{Localization:} Similarly as in \cite{gap}, in order to generate the bounding box of a predicted object from the weight map $\alpha$, we use the simple thresholding technique to segment the heat map, and we take the bounding box that covers the largest connected component in the segmentation map. We only do this for the top-1 predicted class. We provide the top-1 results on the ILSVRC validation set in Table~\ref{loc}. We observe that our method GoogLeNet-GWAP yields the lowest localization error of $54.99\%$ on top-1. Furthermore, we observe the outputs of our GoogLeNet-GWAP model with the multi-scaled input image. We have tried three scales: 224, 448 and 672 in this paper. We also provide the localization result by simply average these multi-scale weight maps. With this multi-scale strategy, there is an improvement of $0.9\%$. As shown in Fig.~\ref{fig:maps}, we observe that the proposed method is able to successfully indicate target objects. Different from GoogLeNet-GAP \footnote{We download the pre-trained model from https://github.com/metalbubble/CAM}, our method tends to localize the full extent of objects and provide shape and contour information about objects. The larger resolution of weight maps, the more precise regions of objects emerge in. These results are promising since we did not use any annotated bounding boxes.   

\section{Semi and Weakly Supervised Detection}
\label{exp:swd}

In this section, we evaluate our method on the PASCAL VOC benchmark. Same as in~\cite{dai2016r}, we train the proposed model in Section~\ref{sec:swd} on the union set of VOC 2007 trainval and VOC 2012 trainval, and evaluate on VOC 2007 test set. Object detection accuracy is measured by mean Average Precision (mAP). For each category in the training set, some of the images have both the image-level labels and the ground truth bounding boxes, while other images only have image-level labels. For comparison, we use the same training settings in~\cite{dai2016r}. We use ResNet-50 as the basic model. The weakly supervised task and the supervised detection task are end-to-end trained jointly. The weight for the loss of the weakly supervised task is $0.1$ in all these experiments. 

\begin{figure*}[t]
	\centering
	\subfigure[Horse\&Person]{ 
		\includegraphics[width=0.2\linewidth]{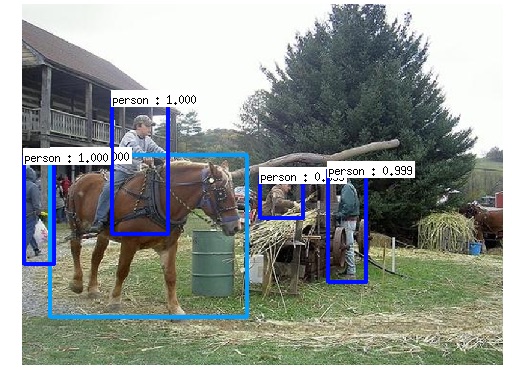}}
	\subfigure[Horse]{ 
		\includegraphics[width=0.2\linewidth]{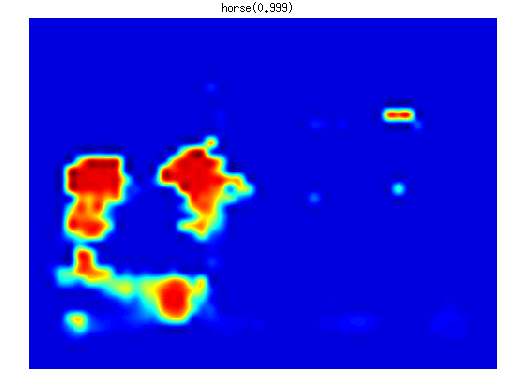}}
	\subfigure[Person]{ 
		\includegraphics[width=0.2\linewidth]{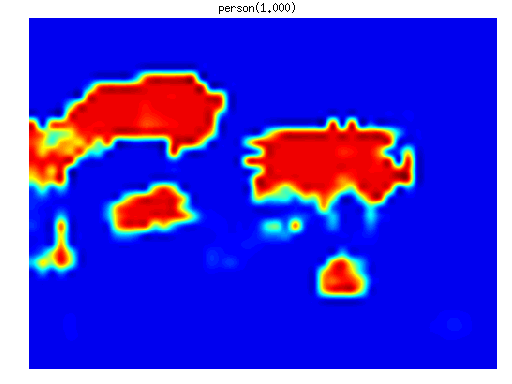}} \\
	\subfigure[Cat\&Dog]{ 
		\includegraphics[width=0.2\linewidth]{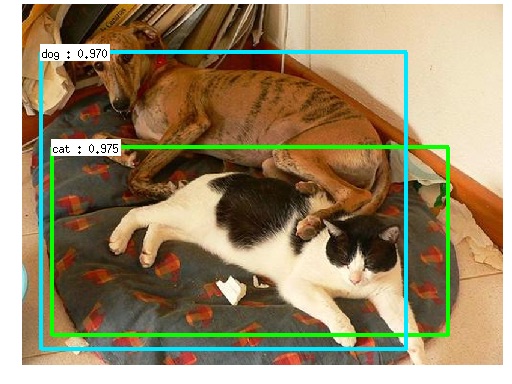}}
	\subfigure[Cat]{ 
		\includegraphics[width=0.2\linewidth]{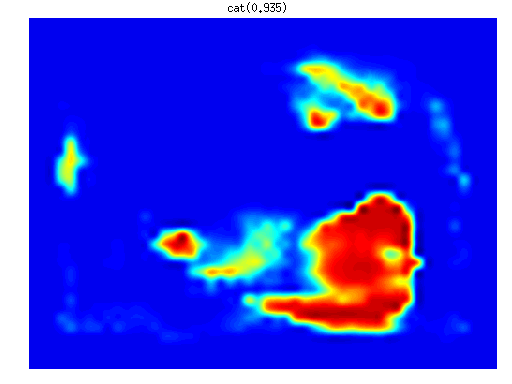}}
	\subfigure[Dog]{ 
		\includegraphics[width=0.2\linewidth]{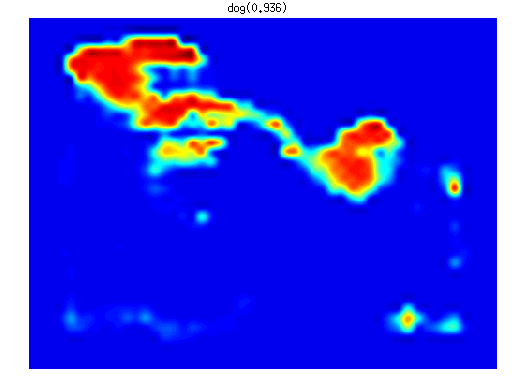}}  \\
	\subfigure[Bus\&Car\&Person]{ 
		\includegraphics[width=0.2\linewidth]{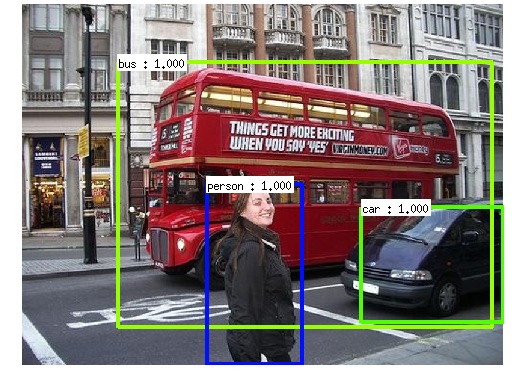}}
	\subfigure[Bus]{ 
		\includegraphics[width=0.2\linewidth]{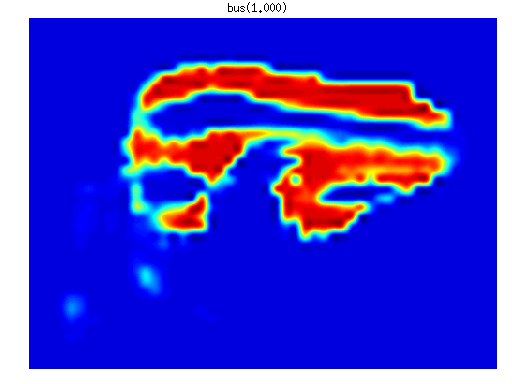}}
	\subfigure[Car]{ 
		\includegraphics[width=0.2\linewidth]{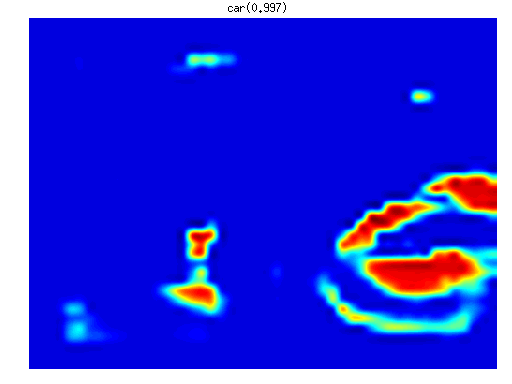}}
	\subfigure[Person]{ 
		\includegraphics[width=0.2\linewidth]{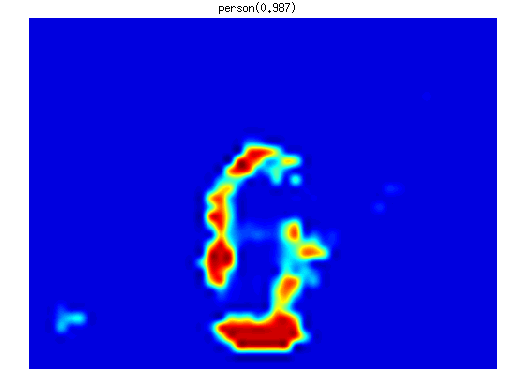}} \\
	\caption{Visulization of some detection results and the corresponding class-specific weight maps.}
	\label{fig:swd}
\end{figure*}

Results are shown in Table~\ref{table:swd}. The baseline model is the standard R-FCN~\cite{dai2016r} detection model that can only be trained with images that have ground truth bounding boxes. We combined GWAP and GAP method with R-FCN. Then the images with image-level labels are also used for training. We show that training with few labeled bounding boxes and large-scale image-level labels significantly improve the performances. This demonstrates that the multi-task framework efficiently regularizes the learning when lack of supervised training data for detection. We also show that the proposed GWAP method performs better than GAP for this task. We report some visualizations of the class-specific weight map in Fig.~\ref{fig:swd}. 

\begin{table}[t] \small
	\caption{Comparisions on PASCAL VOC 2007 test set with different proportions of images that have ground truth bounding boxes. (mAP($\%$))}
	\label{table:swd}
	\centering
	\begin{tabular}{l|cccc}
		\hline \hline
		\diagbox{Method}{Proportion} & 0.01 & 0.02 & 0.05 & 0.1  \\
		\hline
		R-FCN & 29.89 & 40.25 & 53,24 & 61.05  \\
		R-FCN + GAP & 39.10 & 45.85 & 57.36 & 62.32 \\
		R-FCN + GWAP (ours) & 39.86 & \textbf{47.53} & 57.34 & \textbf{63.17} \\		
		R-FCN + GWAP + GAP (ours) & \textbf{40.24} & 47.20 & \textbf{57.48} & 62.55 \\
		\hline
	\end{tabular}
\end{table}

\section{Conclusion}

In this paper, we revisit the global weighted average pooling (GWAP) method and develop the class-agnostic/specific GWAP modules for simultaneous pixel-level localization and image-level classification with only image-level labels for training.
We show that precise regions of objects can be obtained by the proposed methods without using supervised annotations. We further propose a multi-task framework that combines our class-specific GWAP module with R-FCN. We show that this framework can use the data with only image-level labels to significantly improve the generalization of the object detection model. We hope that the results of this paper will encourage future exploration in weakly supervised learning and object detection with convolutional neural networks. We also expect the GWAP module to have useful applications. In the future, we plan to combine our method with the other weakly supervised detection methods.

\section*{References}
\bibliography{egbib}

\end{document}